\documentclass{article}
\usepackage{bm}
\usepackage{float}

\usepackage[preprint]{neurips_2024}

\usepackage[utf8]{inputenc} % allow utf-8 input
\usepackage[T1]{fontenc}    % use 8-bit T1 fonts
\usepackage{hyperref}       % hyperlinks
\usepackage{url}            % simple URL typesetting
\usepackage{booktabs}       % professional-quality tables
\usepackage{amsfonts}       % blackboard math symbols
\usepackage{amsmath}        % for \text and advanced math formatting
\usepackage{graphicx}       % for including graphics
\usepackage{nicefrac}       % compact symbols for 1/2, etc.
\usepackage{microtype}      % microtypography
\usepackage[table,dvipsnames,svgnames]{xcolor}

\usepackage{ulem}
\usepackage{tcolorbox}
\usepackage{placeins} % Add this in the preamble
\usepackage{cleveref}
\usepackage{wrapfig}
\usepackage{enumitem}
\usepackage{listings}
\usepackage[utf8]{inputenc}
\usepackage{booktabs}
\usepackage{colortbl}
\usepackage[table]{xcolor}

\usepackage{multirow}

\usepackage{comment}

\newcommand{\reffig}[1]{\text{Figure~\ref{#1}}}
\newcommand{\reftab}[1]{\text{Table~\ref{#1}}}

\newcommand{\refsec}[1]{\text{Section~\ref{#1}}}

\makeatletter
\renewcommand{\@fnsymbol}[1]{\@arabic{#1}}
\makeatother

\title{NeurIPS 2025 E2LM Competition :  Early Training  Evaluation of Language Models}

\author{
Mouadh Yagoubi\thanks{Technology Innovation Institute (TII), UAE} \quad
Yasser Dahou\footnotemark[1] \quad
Billel Mokeddem\footnotemark[1] \quad
Younes Belkada\footnotemark[1] \\
\textbf{Phuc H. Le-Khac}\footnotemark[1] \quad
\textbf{Basma El Amel Boussaha}\footnotemark[1] \quad
\textbf{Reda Alami}\footnotemark[1] \quad
\textbf{Jingwei Zuo}\footnotemark[1] \\
\textbf{Damiano Marsili}\thanks{California Institute of Technology, Pasadena, CA, USA} \quad
\textbf{Mugariya Farooq}\footnotemark[1] \quad
\textbf{Mounia Lalmas}\thanks{Spotify, UK} \\
\textbf{Georgia Gkioxari}\footnotemark[2] \quad
\textbf{Patrick Gallinari}\thanks{ISIR - Sorbonne University, France} \quad
\textbf{Philip Torr}\thanks{University of Oxford, UK} \quad
\textbf{Hakim Hacid}\footnotemark[1]
}

\date{}

\begin{document}

\maketitle

\begin{abstract}
Existing benchmarks have proven effective for assessing the performance of fully trained large language models. However, we find striking differences in the  early training stages of small models, where benchmarks often fail to provide meaningful or discriminative signals. To explore how these differences arise, this competition tackles the challenge of designing scientific knowledge evaluation tasks specifically tailored for measuring early training progress of language models. Participants are invited to develop novel evaluation methodologies or adapt existing benchmarks to better capture performance differences among language models.
To support this effort, we provide three pre-trained small models (0.5B, 1B, and 3B parameters), along with intermediate checkpoints sampled during training up to 200B tokens. All experiments and development work can be run on widely available free cloud-based GPU platforms, making participation accessible to researchers with limited computational resources.
Submissions will be evaluated based on three criteria: the quality of the performance signal they produce, the consistency of model rankings at 1 trillion tokens of training, and their relevance to the scientific knowledge domain.
By promoting the design of tailored evaluation strategies for early training, this competition aims to attract a broad range of participants from various disciplines, including those who may not be machine learning experts or have access to dedicated GPU resources. Ultimately, this initiative seeks to make foundational LLM research more systematic and benchmark-informed from the earliest phases of model development.

\end{abstract}

\subsection*{Keywords}
Early training analysis, Evaluation benchmarks, Scientific knowledge, Low-resource ML research, Language Models.  
\\
\\

\section{Competition description}

\subsection{Background and impact}
\label{sec:background}

Large Language Models (LLMs) have advanced at an unprecedented pace in recent years~\cite{grattafiori2024llama,yang2024qwen2,malartic2024falcon2,zhao2023survey}. In response, the research community has increasingly focused on developing benchmarking frameworks to assess their capabilities in a variety of domains, including language, mathematics, and reasoning. Early evaluations primarily focused on measuring the language modeling skills \cite{zellers2019hellaswag, sakaguchi2021winogrande}, a fundamental competence that appears to be increasingly well-acquired by newer LLM variants saturating such benchmarks. Recently, attention has shifted towards evaluating scientific knowledge, code generation and reasoning capabilities, using benchmarks such as MMLU \cite{hendryckstest2021} and its  enhanced version MMLU-pro \cite{wang2024mmlu}, GPQA \cite{rein2024gpqa}, GSM8K \cite{cobbe2021training},  MATH \cite{hendrycks2021measuring}, and LiveCodeBench \cite{jain2024livecodebench}. A common theme across these benchmarks is the increasing complexity of tasks, aimed at testing the limits of frontier models.

Despite these advances, Small Language Models (SLMs) with lower than 7B parameters consistently underperform relative to their larger counterparts on these challenging benchmarks. This performance gap is particularly problematic when training SLMs from scratch, as these benchmarks fail to offer  discriminative signals during early training phases (typically between 0-200 billion tokens). Consequently, they provide limited guidance for training dynamics regarding: hyperparameter search, data mixture and architectural sweeps. Developing more granular and targeted evaluation benchmarks in the scientific knowledge domain would enable more systematic investigation of fundamental research questions during the early training stages of SLMs.

\begin{figure}[t!]
    \centering
    \includegraphics[width=\textwidth, height=0.2\textheight]{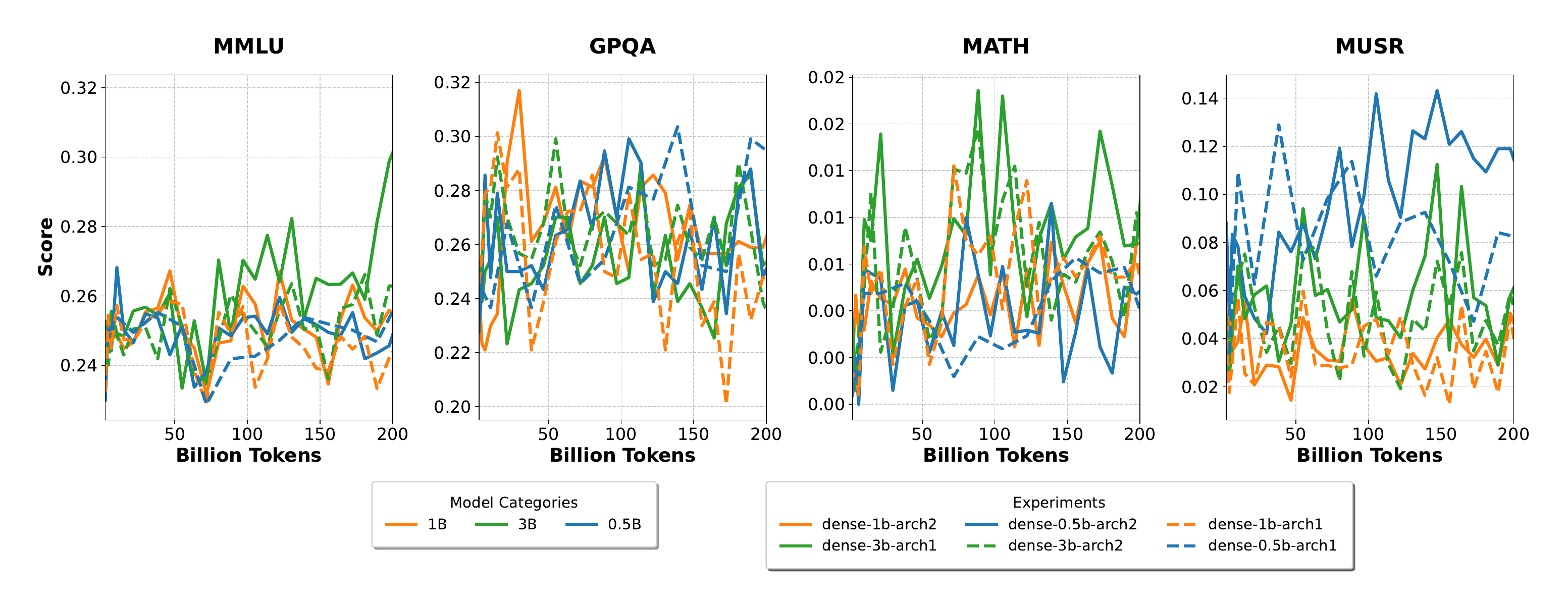}
    \caption{ Noisy results obtained with state-of-the-art benchmarks with different models sizes.}
    \label{fig:noise}
\end{figure}

Concretely, we provide three pre-trained SLMs (0.5B, 1B, and 3B) along with intermediate checkpoints up to 200 billion tokens. As shown in \reffig{fig:noise}, current benchmarks are often noisy and fail to provide a clear signal to measure progress, raising several fundamental questions: Are SLMs not learning any scientific knowledge early in the training? Or are the benchmark questions too difficult at this stage? Is the noise due to a lack of acquired knowledge, or could it be related to how the questions are formatted from a linguistic standpoint? 
Are there ways to refine current benchmarks to better extract meaningful signals? 

The primary goal of this competition is to investigate these questions by inviting participants to design and develop novel evaluation tasks that enable a progressive assessment of scientific knowledge in SLMs, hence, enabling a progressive assessment of scientific knowledge in SLMs and producing more informative performance signals. Specifically, the expected impact of the challenge can be summarized as follows:

\begin{itemize}[leftmargin=*, itemsep=0pt, topsep=0pt]
\item \textbf{Enhancing SLMs research:} By enabling more rigorous and conclusive studies on smaller models, this competition will open opportunities to develop stronger SLMs in scientific knowledge, code, and mathematics--ultimately helping to bridge the gap with larger models through meaningful analysis across a large number of small-scale experiments.

%\item \textbf{Transfer to large models:} motivated by potential lessons from SLMs, better decisions will be taken when training larger models with regards to: data mixtures, hyper-parameters, and architecture. We refer the reader to \cite{chen2024role,yang2022tensor} for a comprehensive survey on the role of small models in enhancing LLMs.

\item \textbf{Transfer to large models:} Lessons learned from studying SLMs can help inform better decisions when training larger models, particularly in terms of data mixtures, hyper-parameters, and architectural choices. We refer the reader to~\cite{chen2024role,yang2022tensor} for a comprehensive overview of the role of small models in advancing LLMs.

%\item \textbf{Understanding the learning dynamics:} the notion of complexity and convergence is not apparent in the current benchmarks. Intuitively, benchmark complexity should be adapted to the models convergence. Hence, solving these questions will uncover the ways in which these models evolve.

\item \textbf{Understanding the learning dynamics:} Current benchmarks often do not fully reflect the convergence or complexity of the model in a meaningful way. Benchmark difficulty should ideally evolve with a model’s level of training. Tackling this problem may offer insights into how models learn over time.

%\item \textbf{Expand the reach of LLM research:} we aim to encourage contributions from researchers beyond the computer science community, inviting ideas and benchmarks drawn from their respective domains of expertise. This cross-disciplinary collaboration has the potential to broaden the scope and enhance the impact of LLM research.

\item \textbf{Expand the reach of LLM research:} We aim to encourage participation from researchers beyond the core machine learning community, inviting them to bring evaluation ideas and benchmarks rooted in their domain expertise. Such interdisciplinary collaboration has the potential to broaden the impact and relevance of LLM research.

\end{itemize}

To this end, participants can either adapt existing benchmarks or design entirely new tasks from scratch. Moreover, we provide a baseline approach that has been successfully tested on the provided checkpoints (see \refsec{sec:tasks}). This baseline draws inspiration from the work of \cite{muennighoff2024olmoe}, who proposed a variant of the well-known MMLU benchmark \cite{hendryckstest2021}, referred to as MMLU-var. In this variant, prompts are reformulated in a completion/close-style format rather than as multiple-choice questions. The motivation for this adaptation stems from the observation that LLMs typically gain proficiency in multiple-choice questions only in later training stages (e.g., after 1 trillion tokens), which already represents a substantial computational budget.  

Furthermore, the results on MMLU-var (\reffig{fig:mmlu-comparison}) demonstrate how simple prompt modifications (see Appendix \ref{prompt-mmlu}) can substantially improve the evaluation of SLMs. Grade-school-level benchmarks like ARC-easy \cite{allenai:arc} and SciQ \cite{SciQ} also provide useful signals (refer to Appendix \ref{app-all-benchmarks-plots}). That said, benchmark complexity is inherently multi-dimensional; while MMLU-var addresses linguistic complexity in prompts, participants are encouraged to explore additional dimensions such as semantic depth and domain-specific reasoning in areas like mathematics or code generation. 

\subsection{Novelty}

Research on LLMs has been rapidly evolving across multiple fronts, including how to evaluate them effectively~\cite{weidinger2025toward}, how to improve their efficiency\cite{wan2023efficient}, their training scaling laws\cite{kaplan2020scaling,hoffmann2022training}, and how to combine or adapt them for diverse use cases\cite{hadi2023survey}. In recent NeurIPS editions, several competitions have been introduced to address the high costs associated with training and deploying them: the LLM Merging Challenge explored how to combine existing models to create more powerful ones without additional training~\cite{tam2024llm}, while the Edge LLM Challenge~\cite{liu2024edge} focused on developing efficient, optimized models capable of running on resource-constrained edge devices. A Single-GPU Fine-Tuning and Inference Challenge  was also organized to promote the development of methods that reduce the computational costs of using LLMs in practice~\cite{llmefficiency}. 

While these challenges target fully trained model's applicability, our proposal shifts the focus towards SLMs and explores how best to evaluate their acquisition of scientific knowledge during the early phases of training. Drawing from foundational work in cognitive development and education~\cite{fischer1985stages,marzano2006new}, we are inspired by the idea that evaluating learning requires sensitivity to both developmental stage and type of understanding. For example, \cite{fischer1985stages} emphasized that learners acquire increasingly abstract forms of reasoning over time, suggesting that evaluation tasks should be developmentally aligned—not just more difficult, but also structured differently at each stage. Similarly, taxonomies in \cite{marzano2006new,krathwohl1964taxonomy}  propose that comprehension, analysis, and knowledge application emerge progressively.

By analogy, we argue that evaluation strategies for SLMs should not focus solely on final performance, but rather take into account the learning dynamics and training trajectory. Just as students are not assessed on differential equations before mastering algebra, language models should be evaluated on tasks aligned with their stage of development. We argue that early-stage benchmarks should probe intermediate reasoning capabilities and the gradual acquisition of scientific knowledge in a way that reflects how understanding evolves with exposure and training.

This competition aims to provide the research community with new evaluation benchmarks that reflect the learning dynamics of Small Language Models. It will equip participants with tailored tools to assess SLMs, with a particular emphasis on reasoning and scientific knowledge domains. 

%While much of the current focus has been on designing complex benchmarks for advanced LLMs, we argue that evaluation strategies must be adapted to suit the specific characteristics of each model family. In this context, SLMs constitute a distinct and important subdomain within the broader LLM ecosystem—offering both computational accessibility and unique developmental trajectories.

By establishing dedicated reasoning- and scientific-knowledge-focused evaluation benchmarks for SLMs, we aim not only to improve the evaluation of SLMs but also to support broader LLM research. Fine-grained evaluation at a small scale enables more informative experimentation, potentially improving decisions in large-scale trainings. Ultimately, we believe that addressing the challenge of evaluating SLMs in a competition setting will foster innovation in benchmark design and inform the development of metrics better suited to the evolving needs of the research and industry communities.

%Indicate whether this is an entirely new competition, or part of a series, eventually re-using old data. 
%If this is an updated version of a previously accepted competition (either at NeurIPS or another venue), extensively comment on the updates you introduced and provide justification if the competition is substantially similar to previous iterations. 
%If you are aware of similar previous competitions, describe the key differences with your proposal.
\subsection{Data} \label{sec:data}

To support participants and facilitate the validation of their benchmarks, we trained multiple models using two distinct data mixtures. Below, we provide a detailed description of each:
\begin{itemize}[leftmargin=*, itemsep=0pt, topsep=0pt]
 \item \textbf{Web-only Dataset}: This consists of a random subset of FineWeb \cite{penedo2024the}, a cleaned and deduplicated English web corpus derived from CommonCrawl\footnote{\url{https://commoncrawl.org}}. Models trained on this mixture typically acquire broad linguistic fluency and general world knowledge.
 \item \textbf{Scientific Knowledge Data Mixture:} This mixture comprises selected subsets from the following publicly available datasets: FineWeb-edu \cite{lozhkov2024fineweb-edu} (50\%), The Stack V1 \cite{Kocetkov2022TheStack} and The Stack V2 \cite{lozhkov2024starcoder} (21.6\%), InfiMM \cite{han2024infimmwebmath40badvancingmultimodalpretraining} (18.9\%), and TxT360\footnote{All subsets are included except for Common Crawl.} \cite{txt360data2024} (9.5\%). Compared to the Web-only mixture, models trained on this subset are expected to demonstrate stronger performance in logical inference, mathematical problem-solving, and code generation.
\end{itemize}

Each model was trained on 1 trillion tokens using a custom tokenizer with a vocabulary size of 65,000 tokens. To maintain a clear scope for the competition and ensure a fair comparison across submissions, only English-language data was used. Additionally, we provide models at three scales—each instantiated with two distinct architectural variants for the same size: a deep variant (arch1) and a wide variant (arch2), assuming deeper models reason better \cite{allen2023physics}, as shown in Appendix \ref{architectures}. The choice is merely to differentiate models of the same size rather than to compare width vs depth, which is out of the scope of this work. Participants are provided with intermediate checkpoints up to 200 billion tokens, enabling evaluation at various stages of training.

\subsection{Tasks and application scenarios}
\label{sec:tasks}

\begin{figure}[htbp]
    \begin{minipage}{0.65\textwidth}
        \includegraphics[width=\linewidth, height=0.5\linewidth]{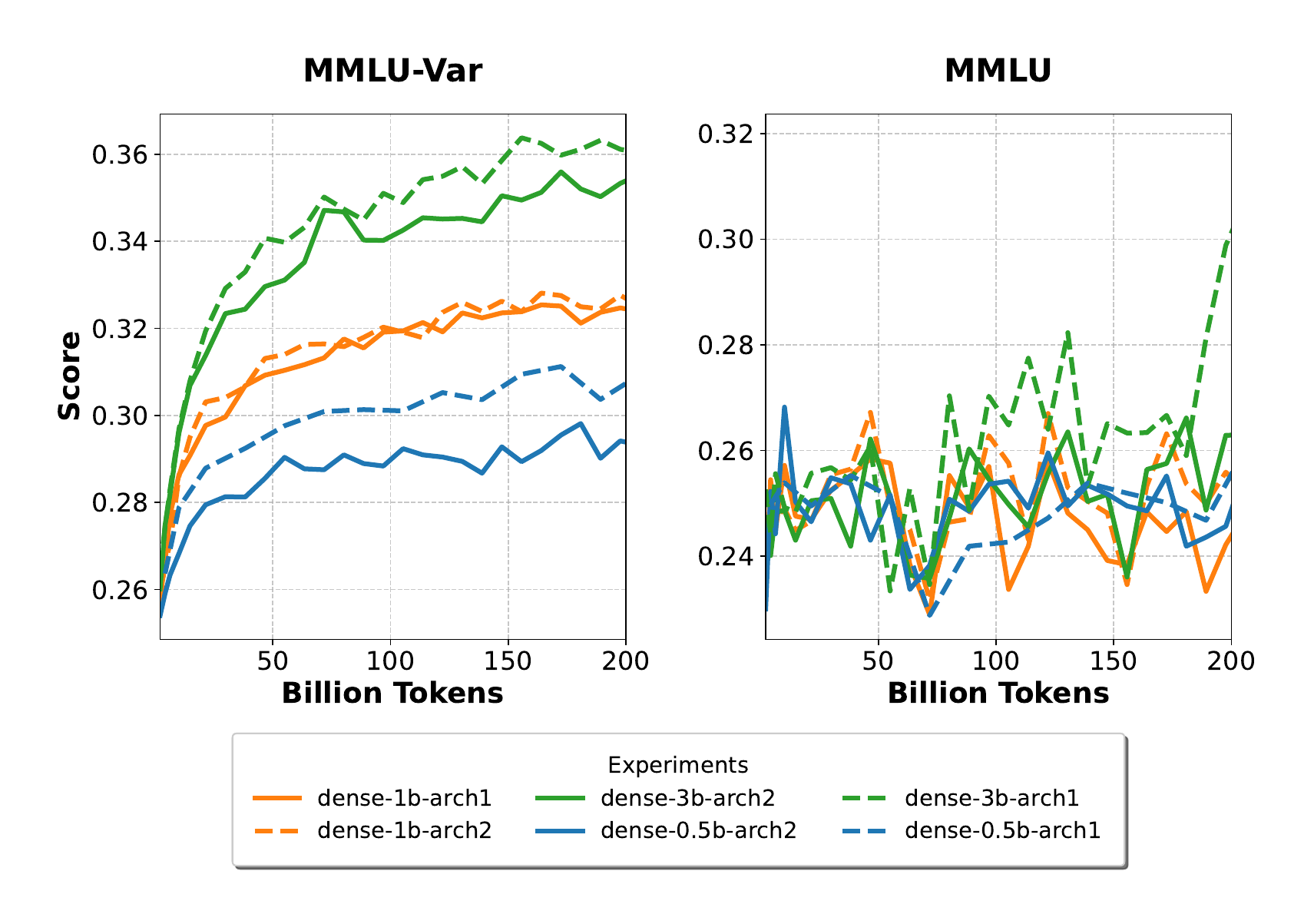}
    \end{minipage}%
    \begin{minipage}{0.3\textwidth}
        \caption{\small \textbf{MMLU-var (left) vs MMLU (right) comparison on different configurations and sizes of models.} Different colors for each size of the model (0.5B, 1B and 3B) and markers distinguish experiments within the same category. MMLU-var allows a clear comparison between variants while MMLU is giving noise (expect for 3B arch2).}
        \label{fig:mmlu-comparison}
    \end{minipage}
\end{figure}

We begin our investigation using the baseline results obtained on the MMLU-var benchmark. ~\reffig{fig:mmlu-comparison} presents a comparison of various model architectures and sizes using both the standard MMLU and its modified variant, MMLU-var~\cite{muennighoff2024olmoe}.

% Solution 1: Using minipages

We observe that results on MMLU do not consistently differentiate between models of varying sizes and architectures (i.e, 0.5B, 1B, and 3B), with no apparent scaling law or pattern, except at later stages of training (around 180 billion Tokens), where the \textit{3B-Arch-2} model clearly outperforms the others. 

In contrast, the MMLU-var (left plot) offers better informative results throughout the training process. On top of ranking models according to size, it also  distinguishes between different architectural variants within the same size. These results suggest that early trained SLMs struggle to explicitly compare and reason over multiple-choices in the input space, but still gives a higher log-likelihood to the correct answer when prompted in a completion manner. Interestingly, we found that tasks requiring general language capabilities (i.e., HellaSwag \cite{zellers2019hellaswag}, WinoGrande\cite{sakaguchi2021winogrande}) provide meaningful signals during early training (0 to 200 BT). Furthermore, a core objective of this competition is to identify and explore similar ideas that reliably evaluate scientific knowledge.

% This highlights MMLU-var's potential as an effective benchmark for evaluating reasoning and scientific knowledge capabilities during pretraining of SLMs.

\begin{figure}[htbp]
\centering
    \begin{minipage}{0.6\textwidth}
        \includegraphics[width=\linewidth, height=0.5\linewidth]{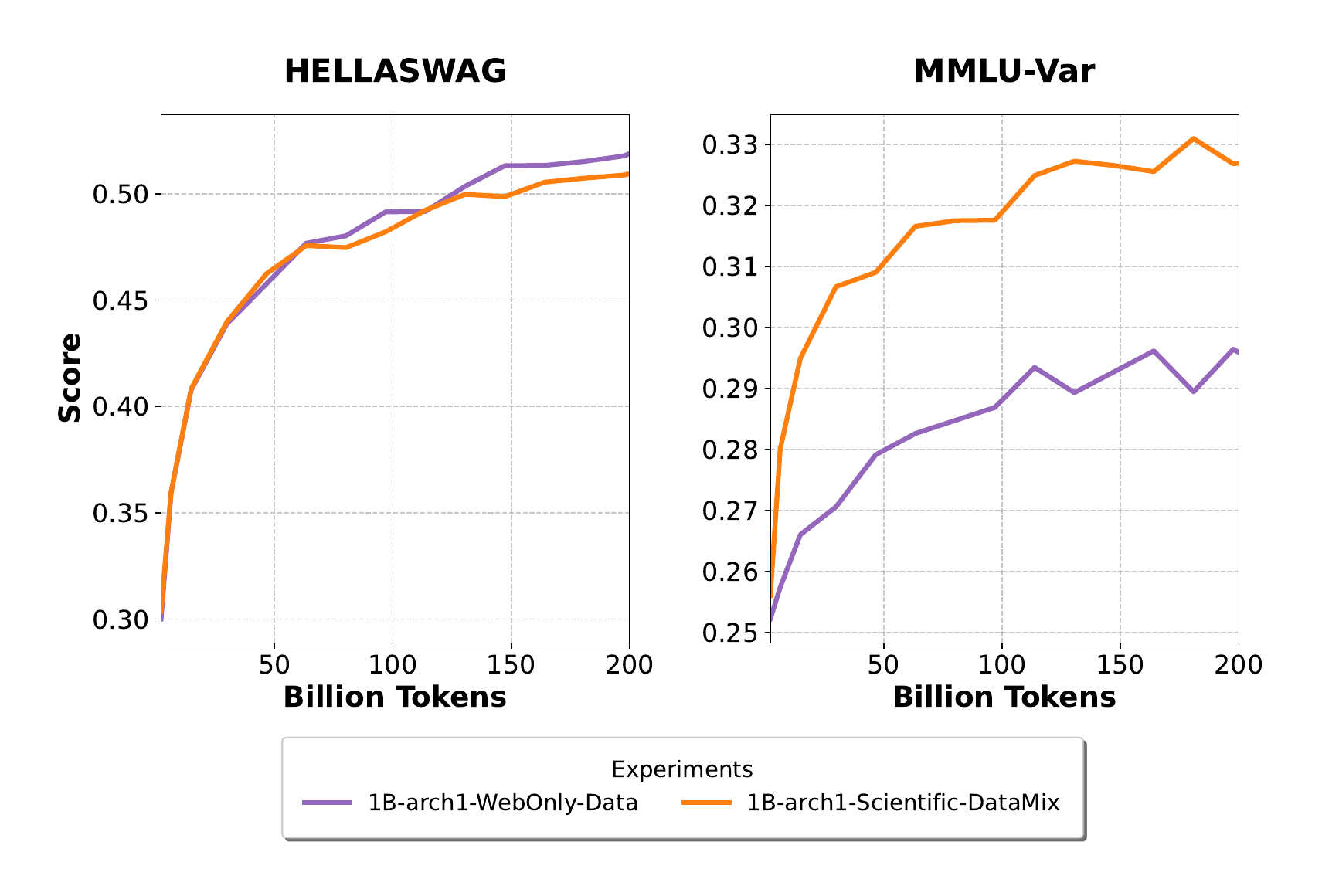}
    \end{minipage}%
    \begin{minipage}{0.3\textwidth}
        \caption{ \small Comparison of results from a 0.5B parameter model trained on two datasets — Web-Only and Knowledge Data-Mixture. While both model configurations achieve similar performance on the HellaSwag benchmark, the model trained on the Knowledge Data-Mixture significantly outperforms the Web-Only model on MMLU-var. %This underscores MMLU-var’s value as a benchmark specifically adapted for evaluating knowledge and reasoning capabilities.
        }
        \label{fig:mmluvar-hellaswag-comparison}
    \end{minipage}
\end{figure}

\textbf{The effect of data mixtures:} we trained a 1B model on a different datasets, one using only web data, and the other using a curated mix of knowledge-rich domains (as described in \refsec{sec:data}).
As shown in \reffig{fig:mmluvar-hellaswag-comparison}, both models perform similarly on HellaSwag. However, on MMLU-var, the model trained on the knowledge-enriched dataset significantly outperforms the web-only variant. This demonstrates that while HellaSwag is effective for evaluating general language understanding in SLMs, it does not adequately capture reasoning or knowledge capabilities. In contrast, MMLU-var clearly differentiates between the two, with the knowledge-trained model achieving higher scores. This eliminates the hypothesis that acquiring scientific knowledge is a fundamental limitation in SLMs' early training and suggests that factors such as data, architecture, or hyperparameters could potentially enhance the model's capacity at this stage.

% Solution 1: Using minipages

% These findings reinforce the idea that while benchmarks like HellaSwag are valuable for tracking language proficiency, they fall short in evaluating reasoning and knowledge. MMLU-var, by comparison, is more sensitive to the presence of structured knowledge in the training data and helps fill a critical gap in current evaluation methodologies.

\subsection{Metrics}
We propose to evaluate submitted solutions based on three main criteria, which will be combined into a global score used for the final ranking: Signal Quality (SQ), Ranking Consistency (RC), and Compliance with Scientific Knowledge Domains (CS).  Additionally, two validation procedures will be systematically applied to all submissions: (i) verification of alignment with established scientific knowledge domains (see Appendix \ref{scientific-alignement check}), and (ii) detection of potential information leakage, specifically the presence of the answer within the question prompt (see Appendix \ref{leakeage-check}).%compliance check will be applied to all submissions to ensure alignment with the scientific knowledge domains (see Appendix E).
 The overall score is computed as a weighted sum:
\begin{equation}
\label{eq:total_score} \text{Score} = \alpha_{1} \times \bm{Score_{\textbf{SQ}}} + \alpha_{2} \times \bm{Score_{\textbf{RC}}} + \alpha_{3} \times \bm{Score_{\textbf{CS}}} 
\end{equation}
Here, $\alpha_{SQ}$,  $\alpha_{RC}$ and $\alpha_{CS}$ are weighting coefficients that reflect the relative importance of each criterion. We set the initial weights as $\alpha_{1} = 0.5$,  $\alpha_{2} = 0.1$ and $\alpha_{3} = 0.4$,  thereby placing greater emphasis on signal quality and compliance to scientific knowledge, which we consider the most important metrics in evaluating submissions.

Participants will be able to compute the signal quality subscore ($\bm{Score_{\textbf{SQ}}}$) locally using the provided model checkpoints (ranging from 0 to 200 BT) along with the accompanying scoring algorithm (provided in a notebook the starting kit). In contrast, the other two subscores cannot be computed independently, as the corresponding checkpoints—from 200 bilion tokens to 1 trilion tokens, as well as the 0.5 billion parameter model trained exclusively on web data—will remain hidden throughout the competition. The global score will, however, be automatically computed upon submission via the Codabench platform, enabling participants to monitor their overall performance. This setup is designed to prevent overly customized solutions tailored specifically to the released checkpoints.  %We present in the following the sub-scores calculation for signal quality and ranking consistency.

\paragraph{Signal Quality metric  (\textit{Score\textsubscript{SQ}})} This score rewards evaluation tasks that produce smooth and informative learning curves throughout training. 
%\textit{Score\textsubscript{SQ}} is composed of two submetrics. 
%
Let $X_{200BT} = \{(i_j, x_j) \mid j = 1, 2, \ldots, n\}$ denote the set of training iterations $i_j$ and corresponding benchmark scores $x_j$ measured up to 200 billion tokens. \textit{Score\textsubscript{SQ}} is composed of two subcomponents: 

\begin{itemize}[leftmargin=*, itemsep=0pt, topsep=0pt]
  \item \textbf{Monotonicity Score}: This component uses Spearman's rank correlation to measure the degree of monotonic improvement over time. Given rank differences $d_j$ between iteration indices and their associated scores, the score is 
  computed as:
\begin{equation}
\label{eq: monotonicity}
    \text{Score}_{\text{Monotonicity}}   = \max\left(0, 1 - \frac{6 \sum d_j^2}{n(n^2 - 1)}  \right)
\end{equation}
%    \[
%    \rho = 1 - \frac{6 \sum d_j^2}{n(n^2 - 1)}
%    \]
    Negative trends yield a score of 0. This formulation ensures that learning curves showing consistent progress receive higher scores.
    %Where $d_j$ is the difference between the ranks of $i_j$ and $x_j$, and $n$ is the number of scores during training.
    %Since Spearman's Rank Correlation yields a negative value for decreasing trends, we assign a score of $0$ to all such cases. Accordingly, we define the monotonicity score, $\text{Score}_{\text{Monotonicity}}$, as follows:
 %   \[
 %   \text{SC} = \text{Score}_{\text{Monotonicity}} = \max\left(0, \rho\right)
 %    \]

 % \item \textbf{Autocorrelation Strength}: Captures temporal self-consistency of the signal, rewarding signals that are coherent over time rather than erratic. For each lag $\ell \in \{1, 2, \ldots, L\}$ with $L = \lfloor n / 4 \rfloor$, we compute the Pearson correlation coefficient between the original signal and its lagged version (with $\bar{x} = \frac{1}{n} \sum_{i=1}^n x_i$):
    \item \textbf{Autocorrelation Strength}: This captures temporal coherence, rewarding signals that are stable over time rather than noisy. For each lag $\ell \in \{1, 2, \ldots, L\}$ with $L = \lfloor n / 4 \rfloor$, we compute the Pearson correlation coefficient between the original score sequence and its lagged version (with $\bar{x} = \frac{1}{n} \sum_{i=1}^n x_i$):
\begin{equation}
\label{eq: autocorr_lag}
  \rho_\ell = \frac{\sum_{i=1}^{n-\ell}(x_i - \bar{x})(x_{i+\ell} - \bar{x})}{\sqrt{\sum_{i=1}^{n-\ell}(x_i - \bar{x})^2} \times \sqrt{\sum_{i=1}^{n-\ell}(x_{i+\ell} - \bar{x})^2}}
\end{equation}
  The autocorrelation score is the average of the absolute correlations across all lags:
 \begin{equation}
\label{eq: autocorr}
 \text{Score}_{\text{AutoCorr}} = \frac{1}{L} \sum_{\ell=1}^L |\rho_\ell|
\end{equation}
\end{itemize}

The final signal quality score (SQ) is a weighted combination of the two components:
\begin{equation}
\label{eq: score_sq}
\text{Score}_{\text{SQ}} = \beta_{\text{1}} \times \text{Score}_{\text{Monotonicity}} + \beta_{\text{2}} \times \text{Score}_{\text{AutoCorr}}
\end{equation}
%
%This dual perspective encourages evaluation tasks that not only reflect model learning in a clear and directional way, but also suppress irrelevant fluctuations.
This dual approach ensures that evaluation tasks reward both directional learning progress and temporal stability, minimizing noisy or erratic signals. Accordingly, we assign equal weight to the two sub-metrics by setting $\beta_{\text{1}}$ to $0.5$ and $\beta_{\text{2}}$ to $0.5$.

\paragraph{Ranking Consistency metric (\textit{Score\textsubscript{RC}})} 
This metric evaluates how well  an evaluation task preserves a consistent ranking of models as training progresses, specifically after processing a large number of tokens (1 trillion tokens). The more stable and reproducible the model rankings over time, the higher the score. Ranking consistency is assessed separately for each model configuration (0.5B, 1B, and 3B), and the final score (\textit{Score\textsubscript{RC}}) is computed as the average across these configurations.

To quantify consistency, we use Kendall’s Tau coefficient \cite{kendall1938}, a widely adopted metric for measuring ordinal correlation between ranked lists. It has been successfully applied in prior work~\cite{kydlicek2024finetasksmultilingualtasks} to evaluate the robustness of model comparisons across training steps. The process proceeds as follows:
\begin{enumerate}[leftmargin=*, itemsep=0pt, topsep=0pt]
    \item {\bf Baseline Ranking (at 200 BT):} For each model size \( s \in \{0.5\text{B}, 1\text{B}, 3\text{B}\} \), we compute the difference between the average performance across the checkpoints between 100 and 200 billion tokens of the two architectures $A\in\{Arch1, Arch2\}$, see the details about the different architectures in \reftab{table:arch}:

\begin{equation}
\begin{aligned}
\text{rank}_{200BT}(s) &= \begin{cases}
1 & \text{if } r(s)>0 \\
0 & \text{otherwise}
\end{cases} \qquad &
\text{r}(s) &= \frac{1}{|K|} \sum_{j \in K} \text{x}^{Arch1}_j(s)-\text{x}^{Arch2}_j(s)
%\label{eq:r}
\end{aligned}
\label{eq:baseline_ranking}
\end{equation}

where $K = \{k \in \mathbb{N} \mid 100 \leq k < 200\}$, and  $\text{x}^{Arch1}_j$ and $\text{x}^{Arch2}_j$ are the scores of models $Arch1$ and $Arch2$ at checkpoint \( j \) respectively. If $Arch1$ is in average better than $Arch2$,  $\text{rank}_{200BT}(s)=1$, Otherwise $\text{rank}_{200BT}(s)=0$. \\

    \item {\bf Ranking Consistency Evaluation (between 200 BT and 1TT):} Let \( P = \{p \in \mathbb{N} \mid 220 \leq p \leq 1000\)\} representing the evaluation points post-200BT.  At each point \( p \in P \), we compare the current model ranking to the baseline.  For each model size \( s \), we define

\begin{equation}
\begin{aligned}
{\text{Score}_{RC}} =  \tau(s) = \frac{1}{|P|} \sum_{p \in P} \tau_p(s) 
\qquad &
\tau_p(s) = \begin{cases}
1 & \text{if } \text{rank}_{200BT}(s) = \text{rank}_p(s)\\
0 & \text{otherwise.}
\end{cases}
\end{aligned}
\label{eq:ranking_consistency}
\end{equation}

where $\text{rank}_p(s)\in\{0,1\}$, and $\text{rank}_p(s)=1$ if $\text{x}^{Arch1}_p(s) > \text{x}^{Arch2}_p(s)$, otherwise $\text{rank}_p(s)=0$.

\end{enumerate}
A higher {Score\textsubscript{RC}} reflects more stable rankings over time, indicating that the evaluation task provides a reliable signal even as training progresses.

\paragraph{Compliance to Reasoning and Knowledge Domains (Score\textsubscript{CS})}
This metric assesses how well the proposed evaluation task aligns with reasoning and scientific knowledge, rather than general language or commonsense capabilities. As illustrated in ~\reffig{fig:mmlu-comparison-all-cs}, tasks designed to probe scientific knowledge (such as MMLU-var) clearly differentiate between models trained on a curated, knowledge-rich dataset and those trained solely on web data. In contrast, benchmarks like HellaSwag, which emphasize commonsense reasoning, tends to yield similar  results across both models types, making them less effective for this competition’s objectives.

To quantify domain compliance,  we compare the average performance of two 1B models across training steps: a model trained on the scientific  knowledge-focused datamixture ($x^{\text{SciKW-DS}}$) and a model trained on web-only data ($x^{\text{Web-DS}}$). The score is define as the normalized performance gap:
\begin{equation}
\text{Score}_{\text{CS}} = \max\left(0, \frac{1}{n} \sum_{i=1}^{n} \left(x_i^{\text{SciKW-DS}} - x_i^{\text{Web-DS}}\right) \right)
\label{eq:compliance_score}
\end{equation}
%
%\begin{equation}
%\text{Score}_{\text{CRK}} = \max\left(0, \frac{\frac{1}{n} \sum_{i=1}^{n} \left(x_i^{\text{KD}} - x_i^{\text{Web}}\right)}{\frac{1}{n} \sum_{i=1}^{n} \left(\frac{x_i^{\text{KD}} + x_i^{\text{Web}}}{2} \right)} \right)
%\end{equation}
%
This formulation ensures a fair comparison across evaluation tasks of varying difficulty levels, rewarding models that are sensitive to knowledge-intensive learning signals.
%To ensure robustness, the score is only computed for tasks where the average score of both models exceeds a minimum threshold (e.g., 20\%).

%\paragraph{Metric Scores Analysis} ~\reftab{tab:benchmark_scores} presents the evaluation metric scores for MMLU-var, MMLU, and HellaSwag. Among the three benchmarks, MMLU-var achieves the highest total score. Its superiority over MMLU stems from two key factors: it provides a stronger learning signal and maintains consistent experiment rankings between 200 billion and 1 trillion training tokens. MMLU-var also outperformed HellaSwag overall, despite the latter scoring higher on the first two metrics. This is because MMLU-var effectively evaluates scientific knowledge, which is the focus of this competition, while HellaSwag primarily targets commonsense reasoning.

\paragraph{Metric Scores Analysis} ~\reftab{tab:benchmark_scores} presents the evaluation scores for the top-performing benchmarks. MMLU-var achieves the highest overall score, followed by ARC-Easy and SciQ, both of which perform well on all metrics. Our analysis suggests that ARC-Easy is particularly effective for assessing scientific knowledge in the early training stages and can serve as a second baseline alongside MMLU-var in this competition. However, in 82\% of SciQ questions, the correct answer appears verbatim in the prompt, likely making the task easier for models. To mitigate this, we will apply a leakage check to all submissions and exclude such questions from benchmark evaluations, as detailed in Appendix \ref{leakeage-check}. Benchmarks targeting scientific knowledge consistently outperform HellaSwag, despite the latter scoring higher on the Consistency and Signal Quality metrics. HellaSwag receives a zero on the Compliance metric, which evaluates whether the benchmark is classified as scientific knowledge (central to this competition). To validate our scoring metric, we ranked all tested benchmarks (see Appendix \ref{fullplots} and \ref{global-score-all} for full results). Although commonsense reasoning tasks such as HellaSwag, Winnogrande and other related benchmarks are included in the full score table, similar submitted benchmarks will not appear on the official leaderboard, as they fail the scientific compliance pre-check. Their scores are reported solely to assess the robustness of our evaluation metric.

%It got a better score than MMLU mainly because it gave a strong signal and its ranking of experiments is consistent between 200 billion tokens and 1 trillion tokens. It also got a better score than HellaSwag because MMLU-var evaluates scientific knowledge while HellaSwag evaluates commune sense knowledge, even if this latter got a better score than MMLU-var in the first two metrics. 

\begin{table}[htbp]
    \begin{minipage}{0.65\textwidth}
        \centering
        \begin{tabular}{l|ccc|c}
            \hline
            \textbf{Benchmark} & \textbf{Score$_\text{SQ}$} & \textbf{Score$_\text{RC}$} & \textbf{Score$_\text{CS}$} & \textbf{Total Score (\%)} \\
            \hline
            MMLU-var & 0.959 & 0.837 & 0.384 & 71.7 \\
            ARC-Easy & 0.832 & 0.822 & 0.394 & 65.5 \\
            SciQ & 0.846 & 0.772 & 0.316 & 62.7 \\
            HellaSwag & 0.992 & 0.936 & 0.000 & 59.0 \\
            GSM8K & 0.655 & 0.915 & 0.369 & 56.6 \\
            \hline
        \end{tabular}
    \end{minipage}%
    \hspace{1cm}% Add horizontal space between table and caption
    \begin{minipage}{0.25\textwidth}
        \caption{ Benchmark scores across different metrics, with Total Score calculated using Equation~\eqref{eq:total_score}.}
        \label{tab:benchmark_scores}
    \end{minipage}
\end{table}
%\vspace{-1.5em}

\subsection{Baselines, code, and material provided}

%Specify what the baseline solution(s) for the competition will be and provide preliminary results.

%Indicate \textbf{how} and \textbf{when} you plan to release the \textit{``starting kit''} for your competition. This should include code for baselines and data loading tools to help the participants easily join the competition. For certain competitions, material provided may include a hardware platform. 

The provided starting kit \footnote{\href{https://gist.github.com/younesbelkada/b0bbf26e53b89af3325ba2118c6f8e26}{%https://github.com/tiiuae/earlytraining-slm-evaluation
Competition starting kit}} includes a set of Jupyter notebooks designed to help participants get started with the competition materials. These notebooks support the development of evaluation tasks by providing access to model checkpoints and allowing prompting and interaction with the provided LLMs. A complete list of available notebooks is included in Appendix \ref{staritg-kit}.
Since the largest model used in the competition has 3 billion parameters, approximately 6GB of GPU memory is sufficient for inference. As a result,   the experiments and development work can be run on a free-tier Google Colab GPU (NVIDIA T4 16GB), ensuring broad accessibility for participants. In addition, to support participants without a strong background in machine learning, a tutorial session will be organized to explain how the lm-evaluation-harness framework works.

\subsection{Website, tutorial and documentation}

The competition website \footnote{\href{https://et-slm-evaluation.github.io/}{https://et-slm-evaluation.github.io/}} will serve as the central hub for all key information, including: (i) competition overview, (ii) rules and terms, (iii) competition timeline and announcements, (iv) links to the starting kit netbooks, (v) tutorial section that will host the competition webinars, and (vi) contact information for the organizers along with a link to the discussion channel. Additionally, we will also host practical webinar sessions focused on best practices for evaluating Large Language Models.
%Include a link to the competition website or a tentative one if not ready yet. The website should be self-contained, presenting all the relevant information about the competition timeline and illustrating the necessary steps to participate. 

%It is strongly suggested to have a \textbf{FAQ/Tutorial section} and a \textbf{dedicated email address} to reach the organizers. These should be highlighted, easily reachable from the home page, made available from the start and regularly updated.
%The website must be online within two weeks after the acceptance notification. The website's entire content should be released as soon as possible and with enough advance on the start of the competition to allow the participants to have time to prepare.

%If available, provide a reference to a published paper or a white paper you wrote describing the problem, and/or explain what tutorial material you will provide.

\section{Organizational aspects}
%The competition will be run on the Codabench platform \cite{xu2022codabench}. The link will be provided on the competition website. Participants will have to: 1) create an account; 2) download a starting kit to prepare their submission; 3) upload on the Codabench platform thier code, respecting   the described interfaces provided in Codabench. Then, the submissions will be used to evaluate the LLM models (via the provided checkpoints), and to compute the global score. The global score will be published on the Codabench competition leaderboard  and the participants will also have access to an additional page with the detailed metrics.

The competition will be hosted on the Codabench platform~\cite{xu2022codabench}, with the link available on the competition website. Participants must: (1) create an account, (2) download the starting kit to prepare their submission, and (3) upload their code to Codabench following the provided interface specifications. Submissions will be used to evaluate LLM models on the provided checkpoints to compute the scores. Finally, the global scores will be regularly displayed on the Codabench leaderboard, along with a detailed metrics page for each submission.
%\subsection{Protocol}

%\textbf{Explain the necessary steps to join the competition:} what will the participants have to do (e.g., download data, create accounts, etc.)? What are the participants supposed to submit (e.g., list of results, code) and where (e.g., upload to a cloud)? How will submissions be evaluated? 

%Also indicate if the competition will consist of several phases, whether you will use a competition platform with online submissions and a leader-board, \textbf{and how you will prevent cheating and overfitting}.  For organizers looking to host on a third-party platform, there are a number of sites and sponsorship opportunities.  \href{https://www.kaggle.com/competitions-research-grants}{Kaggle research grants}, CodaBench, EvalAI, Grand-challenge, and AICrowd platforms have been utilized in prior years.
%Provide your plan to organize beta tests or dry runs of your protocol and/or platform.

\subsection{Rules and Engagement}

%In this section, please provide:
%This challenge will run from 16 June  2025 to 17 October 2025.% Prizes for winning teams are listed in the Prizes section.

\begin{itemize}[leftmargin=*, itemsep=0pt, topsep=0pt]

\item \textbf{Submissions requirement}: each submission must include a set of question–answer pairs and a clearly defined evaluation metric, implemented in Python code, to assess model performance.

% \item  \textbf{Size of submitted solutions}: each submitted benchmark must contain between 100 and 15,000 samples. The upper bound aligns with the size of MMLU test split of 14,042 samples, among the largest state-of-the-art benchmark, allowing participants to build upon or adapt MMLU if desired. We have successfully tested a baseline using MMLU-var, confirming that our infrastructure can handle submissions at this scale. This cap also helps discourage oversized submissions that combine multiple benchmarks unnecessarily. The lower bound of 100 tasks ensures accessibility for individuals or teams with limited resources who wish to explore new ideas and experiment with smaller, focused benchmarks.

\item \textbf{Size of submitted solutions}: submissions must include 100–15,000 samples. The upper limit matches MMLU’s 14,042-sample test split—one of the largest benchmarks—supporting compatibility and discouraging unnecessarily large, combined benchmarks. The 100-sample minimum ensures accessibility for resource-constrained teams exploring focused ideas.

\item \textbf{Competition phases}: this challenge is open to anyone and runs in 3 phases: during phases 1 \& 2, participants can submit their code and view their results on the regularly updated leaderboard; specifically during the warmup phase, the organizers may adjust the global score formula.

% \item \textbf{Additional baselines:} may be released by the organizers to stimulate participation.
\item \textbf{Submission limits:} each participant or team  may submit up to 10 entries per day.
% \item \textbf{Code availability:} Participants are strongly encouraged to make their code publicly available with their submissions.
 \item \textbf{Final ranking:} the final ranking will be based on the global score, calculated by the organizers and shared with all participants. The scoring algorithm will be made available so participants can evaluate their own solutions locally.
\item \textbf{Team accounts:} teams must use a single group account. The use of multiple accounts is not permitted.
 \item \textbf{Prize eligibility:} To be eligible for a prize, a team must open-source its code at least two weeks before the NeurIPS competition workshop.
\end{itemize}

\subsection{Schedule and readiness}

\label{sec:schedule}
% \paragraph{Detailed timeline} 
The competition will consist of three phases to allow for a smooth participation/organization:

In the \textit{Warm-up phase}, participants get familiarized with the provided materials and the proposed format to integrate their evaluation benchmarks. They may submit preliminary solutions and provide feedback to the organizers, which will help refine the competition setup for the next phase.
%They can make initial submissions and offer feedback to organizers. Organizers will use this feedback to adjust and refine the competition setup for the subsequent phase.

In the \textit{Development phase}, participants will test and refine their evaluation tasks using the provided models and checkpoints. They may use their own resources to compute the signal quality score supported by the starting kit. To obtain a global score, participants must submit their benchmarks via the Codabench platform, where evaluations will be conducted using the competition's resources.

%contestants will have the opportunity to test and enhance their solutions by testing them on the provided models. Throughout this phase, participants can use their own resources, leveraging materials from the starting kit. To validate a score, participants must submit their evaluation tasks on the Codabench platform, where they will be re-tested on the models using the competition resources. 

In the \textit{Final phase}, the organizers will validate the final rankings from the development phase. This includes verifying the submitted benchmarks and assessing the robustness of the top-performing solutions through multiple test runs. At the end of this phase, the organizers will also reveal the hidden checkpoints used during the competition to compute the final scores.

The proposed schedule is the following:
\begin{itemize}
\item {\bf  The competition will run from June 16 to October 17, 2025, comprising a 5-week warm-up phase, a 10-week development phase, and a 3-week final phase.} 
\item Announcement of results: 20 October, 2025.
\item Fact sheets and code release by winners due: 22 November, 2025.
\item Presentation of results at NeurIPS competition workshop: 6/7 December 2025.
\end{itemize}

\vspace{-0.3cm}

\subsection{Competition promotion and incentives}

The challenge will be promoted through various channels, including announcements at leading conferences (NeurIPS, ICLR, ICML), relevant mailing lists, and social media platforms. Monetary prizes will be awarded to the top-performing teams: \$6,000 for 1st place, \$4,000 for 2nd place, and \$2,000 for 3rd place. A special award of \$2,000 will also be granted to each of the two best student solutions. A public leaderboard will be kept displaying the top evaluation tasks across a broad range of large language models (LLMs).

%In addition, the authors of the winning solutions will be invited to co-author a joint paper for submission to the NeurIPS 2026 Datasets \& Benchmarks Track.

%Finally, a leaderboard  based on top evaluations tasks across a wide range of LLM models will be   made public. The authors of  winning solutions  will also be invited to write a joint paper to be submitted to the neurips 2026 Datasets \& benchmarks track.

%\newpage
\bibliographystyle{unsrt}
\bibliography{references}

\section{Resources}

\subsection{Organizing team}

%Provide a short biography of all team members, stressing their competence for their assignments in the competition organization. Please note that diversity in the organizing team is encouraged, so please elaborate on this aspect as well.  Make sure to include: coordinators, data providers, platform administrators, baseline method providers, beta testers, and evaluators \textbf{(these info will not count towards the 8 pages limit)}. 

The team proposing this competition is composed of researchers and engineers in the domains of: Deep learning, NLP, data curation and management and software engineering. The short biography of each member is provided below.
The organizers extend their gratitude to \textbf{Isabelle Guyon} for her valuable contributions to the discussions and insights related to the proposed challenge. \\

\noindent {\bf Mouadh Yagoubi (Lead organizer, Baseline method provider, Evaluator)} is Lead Researcher at TII since November 2024. Prior to that, he was a Senior Researcher at IRT SystemX in France, where he  contributed to several collaborative research projects in machine learning and optimization. He received his Ph.D. in Applied Mathematics from INRIA Saclay in 2012. His research interests include machine learning and evolutionary computation, with a particular focus on their application to industrial problems.

\noindent {\bf Yasser Dahou ( Baseline method provider, Data provider, Evaluator)} is a senior researcher at TII, working on multimodal vision-language research. Before joining TII, he earned a PhD in Computer Vision from Dublin City University in 2024, and a Master’s degree in Telecommunication Engineering in Algeria.

\noindent {\bf Billel Mokeddem (Data provider, Evaluator)} is an AI Engineer at TII. Since he joined in February 2024, He worked on distributed text/audio data processing and model-based filtering, and LLMs pre-training. Prior to joining TII, he worked at Qatar Computing Research Institute as a Research Associate from 2021 to 2023, where he mainly focused on Multi-modality and the use of Variational Inference methods in Continuous Reinforcement Learning. He graduated from University of Science and technology Houari Boumediene (USTHB) in Algeria  with a master's degree in Artificial Intelligence in 2021.

\noindent {\bf Younes Belkada (Platform Administrator, Starting kit maintainer, Evaluator)} is a Senior AI Engineer at TII, where he has been working since July 2024. He contributes to various projects involving pre-training, evaluation, and general tooling for AI and LLM-related initiatives. Prior to joining TII, he worked at Hugging Face from 2021 to 2024 as part of the open-source team, helping to develop a range of open-source deep learning libraries for the community. He graduated from ENS Paris-Saclay in 2021 with a master's degree in Machine Learning and Computer Vision.
 
\noindent {\bf Phuc H. Le-Khac (Platform Administrator, Starting Kit maintainer, Evaluator)}
Phuc is a Deep Learning researcher at Technology Innovation Institute in Abu Dhabi and previously obtained his PhD in Representation Learning at Dublin City University.
He is interested in both the technical details as well as the broader theory of learning system. He is currently applying his interest on developing next-generation Multimodal Model.

\noindent {\bf Basma El Amel Boussaha (Data provider, Evaluator)}
is a Lead Researcher at TII, specializing in building Large Language Models (LLMs) for Arabic. She earned her Ph.D. in Natural Language Processing from Nantes University in 2019. Prior to joining TII, she worked at Della AI, where she played a key role in developing AI-driven contract reviewing tools to support legal professionals. Her research continues to bridge cutting-edge AI with real-world impact across languages and domains with a focus on generative AI, question answering, distributed training, and advancing Arabic NLP.

\noindent {\bf Reda Alami (Evaluator)}
is currently a senior researcher at the Technology Innovation Institute in Abu Dhabi. His research interests include finetuning of large language models for reasoning tasks, test time compute scaling and reinforcement learning. He received his Msc degree with first class honours from IMT Atlantique, France.Then, he received his Phd degree from Paris Saclay University, France. He was with Orange Labs, Lannion, France, from 2016 to 2019 and then withTotalEnergies, Saclay, France from 2020 to 2021

\noindent {\bf Jingwei zuo (Evaluator)} Jingwei Zuo is currently a Lead Researcher at the AI Center of the Technology Innovation Institute (TII). He and his team are working on the Falcon Foundation Models, with a focus on exploring next-generation Large Language Models (LLMs) that incorporate emerging architectural designs beyond or in combination with Transformers, as well as optimal training and model scaling strategies. Jingwei received his PhD in Computer Science from Université Paris-Saclay and was awarded the Plateau de Saclay Doctoral Prize for the best scientific production in Information and Communication Science and Technology (ICST) in 2022.
 
\noindent {\bf Damiano Marsili (Evaluator)}
Damiano is a Ph.D student at Caltech advised by Georgia Gkioxari and Pietro Perona. Before his Ph.D, he was an Applied Science Intern at Amazon Robotics working on 3D spatial reasoning. Prior to that, he double-majored at Johns Hopkins in Computer Science and Mathematics.

\noindent {\bf Mugariya Farooq  (Evaluator)}  is an Electronics Engineer with a  Masters in Machine Learning and is currently an AI Engineer in Technology Innovation Institute. She has led evaluations for various models in Falcon family of models with prime focus on optimization and development of model agnostic evaluation pipeline and benchmarks.

\noindent {\bf  Mounia Lalmas (Scientific Advisor)}
Mounia Lalmas is a Senior Director of Research at Spotify, and the Head of Tech Research in Personalisation, where she leads an interdisciplinary team of research scientists, working on personalization. Mounia also holds an honorary professorship at University College London. She also holds an additional appointment as a Distinguished Research Fellow at the University of Amsterdam. Before that, she was a Director of Research at Yahoo, where she led a team of researchers working on advertising quality. She also worked with various teams at Yahoo on topics related to user engagement in the context of news, search, and user-generated content. Prior to this, she held a Microsoft Research/RAEng Research Chair at the School of Computing Science, University of Glasgow. Before that, she was Professor of Information Retrieval at the Department of Computer Science at Queen Mary, University of London. She is regularly a senior programme committee member at conferences such as WSDM, KDD, WWW and SIGIR. She was programme co-chair for SIGIR 2015, WWW 2018 and WSDM 2020, and CIKM 2023.

\noindent{\bf Georgia Gkioxari (Scientific Advisor)}
Georgia is an assistant professor at the Computing + Mathematical Sciences at Caltech.
She obtained her PhD in Electrical Engineering and Computer Science from UC Berkeley, where she was advised by Jitendra Malik. Prior to Berkeley, she earned her diploma from the National Technical University of Athens in Greece. After earning her PhD, she was a research scientist at Meta's FAIR team. In 2021, she received the PAMI Young Researcher Award, which recognizes a young researcher for their distinguished research contribution to computer vision. She is the recipient of the PAMI Mark Everingham Award for the open-source software suite Detectron (2021), the Google Faculty Award (2024) and the Okawa Research Award (2024).
In 2017, Georgia and her co-authors received the Marr Prize for “Mask R-CNN” published and presented at ICCV. She was named one of 30 influential women advancing AI in 2019 by ReWork and was nominated for the Women in AI Awards in 2020 by VentureBeat.

\noindent{\bf Patrick Gallinari (Scientific Advisor)} 
is a professor in Computer Science at Sorbonne University in Paris. His research focuses on statistical learning with applications in different fields such as semantic data analysis and complex data modeling. He discovered the ML domain in the mid 80es when he started to work on Neural Networks, an emerging field at that time. He has been one of the pioneers of this research domain in France/ Europe and worked on NN and on other ML models since that. He investigated different application domains like Information Retrieval, Social Data analysis, User Modeling. Today his main focus is on Physics Aware Deep Learning and on some aspects of Natural Language Processing. He has been leading the Machine Learning team MLIA for some years. He has been director of the computer science lab. at Sorbonne University (LIP6) for 9 years (2005 to 2013) and vice director for 6 years before, He also acted as vice director of the scientific committee of the faculty of engineering at UPMC (2010 to 2021).

\noindent{\bf Philip Torr (Scientific Advisor)}
Professor Philip Torr did his PhD (DPhil) at the Robotics Research Group of the University of Oxford under Professor David Murray of the Active Vision Group. 
He left Oxford to work for six years as a research scientist for Microsoft Research, first in Redmond, USA, in the Vision Technology Group, then in Cambridge founding the vision side of the Machine Learning and Perception Group. He then became a Professor in Computer Vision and Machine Learning at Oxford Brookes University. In 2013, Philip returned to Oxford as full professor where he has established the Torr Vision group.
He won several awards including the Marr prize (the highest honour in vision) in 1998. He is a Royal Society Wolfson Research Merit Award Holder. 
He was elected Fellow of the Royal Academy of Engineering (FREng) in 2019, and Fellow of the Royal Society (FRS) in 2021 for contributions to computer vision. In 2021 he was made Turing AI world leading researcher fellow.

\noindent{\bf 
Hakim Hacid (Scientific Advisor)}
Hakim Hacid is the Chief Researcher of the Artificial Intelligence and Digital Science Research Center in Technology Innovation Institute (TII), leading the diverse efforts around LLM's and Machine Learning. Prior to joining TII, he was an Associate Professor at Zayed University, a customer analytics head at Zain telecom, and a research department head at Bell Labs Research. He is a published author of many research articles in top journals conferences and holds several industrial patents to his credit. His research specialization includes machine learning, databases, natural language processing, security. He obtained his PhD in Data Mining/Databases and also a double master's in Computer Sc (Master by Research \& Professional Master) from University of Lyon, France.

\subsection{Resources provided by organizers}
In order to evaluate submissions and compute the global score, a dozen H100 GPUs will be made available through GCP infrastructure to power the compute workers connected to Codabench. In addition to a webinar that will be organized to help participants get started, a training session on lm-eval-harness framework will also be provided and made available on the competition website to support participants who are not experts in machine learning.

\subsection{Support requested}
We would greatly appreciate the support of the NeurIPS 2025 Competition Track organizers, particularly in promoting our challenge through their official channels to help us reach participants worldwide.

\newpage
\appendix

\section{Results of state-of-the-art benchmarks with different model sizes}
we provide in this section the results of state-of-the-art benchmarks
\label{app-all-benchmarks-plots}
\begin{figure}[h]
    \centering
    \includegraphics[width=\linewidth]{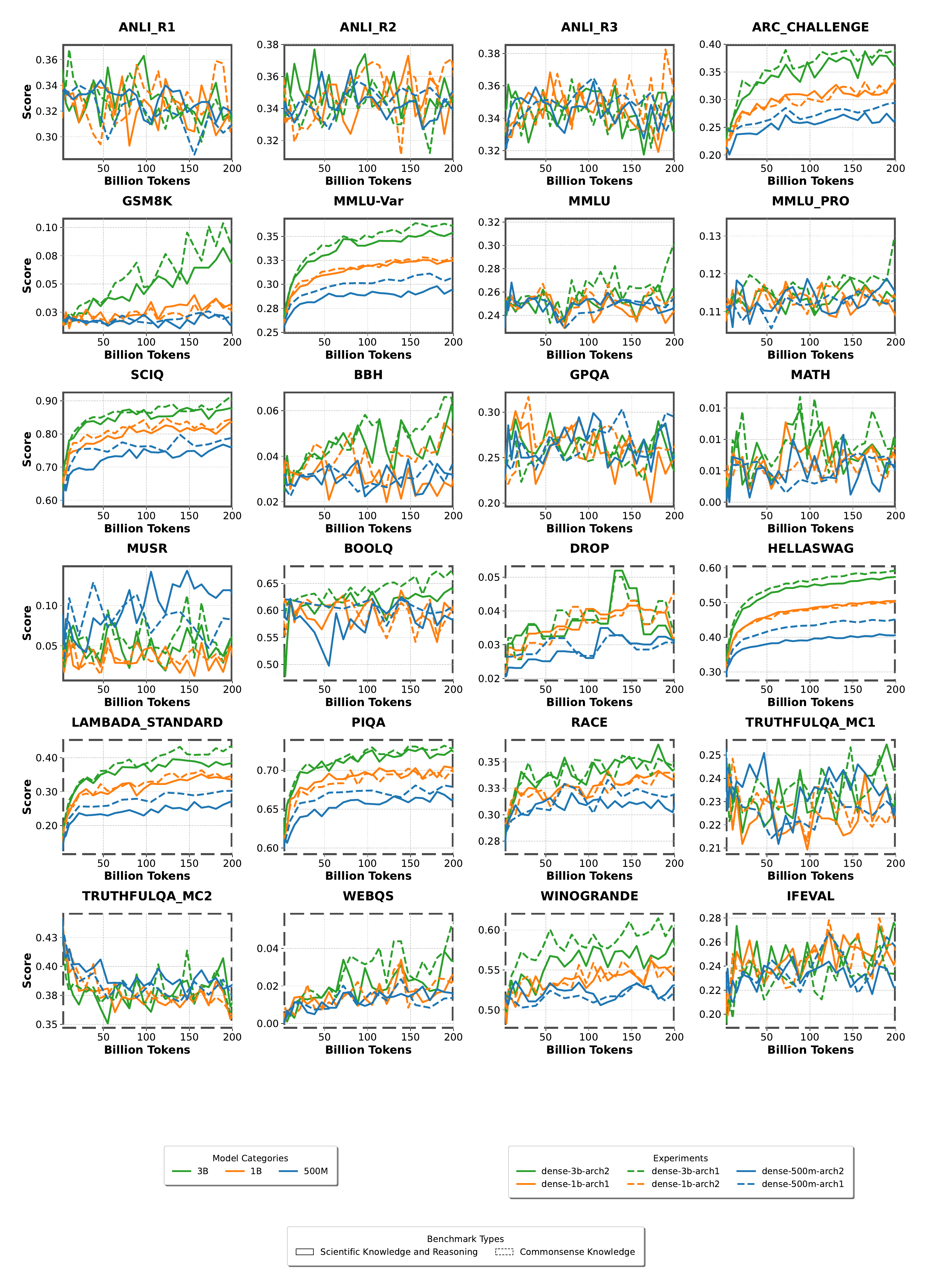}
    \caption{Results of state-of-the-art benchmarks with different models sizes.}
    \label{fig:enter-label}
\end{figure}

\newpage
\section{Example of MMLU and MMLU-var benchmarks}
\label{prompt-mmlu}
We present in this section an example of an evaluation task to highlight the differences between the standard version of MMLU and the MMLU-var variant.

The \textbf{standard MMLU benchmark} uses a multiple choice format: The model receives a question along with a list of possible answers (A, B, C, D) and is prompted to select the correct one. For example, in a question about k-means clustering, the model sees both the question and all four answer choices in the input.

\begin{quote}
\textbf{Standard MMLU prompt example:} \\
\texttt{Question: You want to cluster 7 points into 3 clusters using the k-Means Clustering algorithm. Suppose after the first iteration, clusters C1, C2 and C3 contain the following two-dimensional points...} \\
\texttt{A. C1: (3,3), C2: (4,4), C3: (6,6)} \\
\texttt{B. C1: (3,3), C2: (6,6), C3: (12,12)} \\
\texttt{C. C1: (6,6), C2: (12,12), C3: (12,12)} \\
\texttt{D. C1: (0,0), C2: (48,48), C3: (35,35)} \\
\texttt{Answer:}
\end{quote}

In contrast, \textbf{MMLU-var} uses a continuation-based approach. The model is given only the question followed by the prompt \texttt{"Answer:"}, without any answer choices. The system then evaluates the likelihood the model assigns to each possible answer (as separate completions), and selects the one with the highest log-probability.

\begin{quote}
\textbf{MMLU-var prompt example:} \\
\texttt{Question: You want to cluster 7 points into 3 clusters using the k-Means Clustering algorithm. Suppose after the first iteration, clusters C1, C2 and C3 contain the following two-dimensional points...} \\
\texttt{Answer:}
\end{quote}

Each answer candidate (e.g., \texttt{C1: (3,3), C2: (4,4), C3: (6,6)}) is then evaluated separately by computing its log-probability as a continuation of the prompt above.

\section{Models architectural details}
\label{architectures}
\paragraph{}

\begin{center}
\begin{table*}[ht!]
\centering
\begin{tabular}{|c|c|c|c|c|c|c|}
\hline
Model size & Model name & \# layers & \# attention heads & Hidden size & Intermediate size  \\
\hline
\multirow{3}{*}{0.5B} & dense-0.5B-arch1 & 32 & 16 & 1024 & 3072 \\
& dense-0.5B-arch2 & 16 & 20 & 1280 & 3072 \\
%& dense-500m-arch1-WebOnly & 32 & 16 & 1024 & 3072 \\
%& dense-500m-arch1-bad & 32 & 16 & 1024 & 3072 \\
%& dense-500m-arch2-bad & 10 & 24 & 1536 & 5376 \\
%& dense-500m-arch3-bad & 5 & 16 & 2048 & 6144 \\
%& dense-500m-arch4-bad & 3 & 16 & 2048 & 10240 \\
\hline
\multirow{3}{*}{1B} & dense-1b-arch1 & 32 & 24 & 1536 & 4096 \\
%& dense-1b-arch1-fix-bad & 32 & 20 & 1280 & 6144 \\
& dense-1b-arch2 & 16 & 16 & 2048 & 7168 \\
\hline
\multirow{3}{*}{3B} & dense-3b-arch1 & 32 & 20 & 2560 & 8960 \\
& dense-3b-arch2 & 16 & 24 & 3072 & 15360 \\
\hline
\end{tabular}
\caption{Model architectures. kv heads = 4 for all the models.}
\label{table:arch}
\end{table*}

\end{center}
\newpage
\section{Starting kit description}
\label{staritg-kit}

%TODO: add few comments about the fact that the notebooks can be run on Google Colab for free and can be extensive to Kaggle Multi-GPU

Since the maximum model size we offer is around 3 Billion parameters, all starting kits can be executed on free-tier devices that can be accessible world-wide such as Google Colab or Kaggle free-tier GPU notebooks. This makes the competition free from compute constraints. 

\begin{itemize}
    \item \verb|0-Basic_Competition_Information|: This notebook contains general information concerning the competition organization, phases, deadlines and terms. The content is the same as the one shared in the competition Codabench page.
    
    \item \verb|1-How_to_interact_with_model|: This notebook aims to familiarize participants with the tools that are used to interact with the model and perform some easy text generation tasks. 
    
    \item \verb|2-How_to_evaluate_a_model|: Shows participants how a checkpoint of the model can be evaluated using lm-evaluation-harness package.
    
    \item \verb|3-Reproduce_baseline_results|: his notebook shows how the baseline results (MMLU-Var on a single checkpoint) could be reproduced. It includes integrating the MMLU-Var benchmark within the lm-evaluation-harness package and running it to get its result.

    \item \verb|4-How_to_Contribute|:  This notebook explains how to fully integrate a new task within the the lm-evaluation-harness package.

    \item \verb|5-Scoring|: This notebook shows firstly how the score is computed by describing its different components. Next, it provides a script which can be used locally by the participants to obtain a score for their contributions. We encourage participants to evaluate their solutions via codabench (which uses the same scoring module as the one described in this notebook).
     \item \verb|6-Submission_examples|: This notebook presents the composition of a submission bundle for Codabench and usable parameters
    \\
    \\
\end{itemize}
\newpage
\section{Metrics Calculation Examples}
\label{fullplots}
\subsection[ScoreSQ Calculation examples]{$Score_{SQ}$ Calculation examples}

\begin{figure}[htbp]
    \centering
    \includegraphics[width=0.9\textwidth]{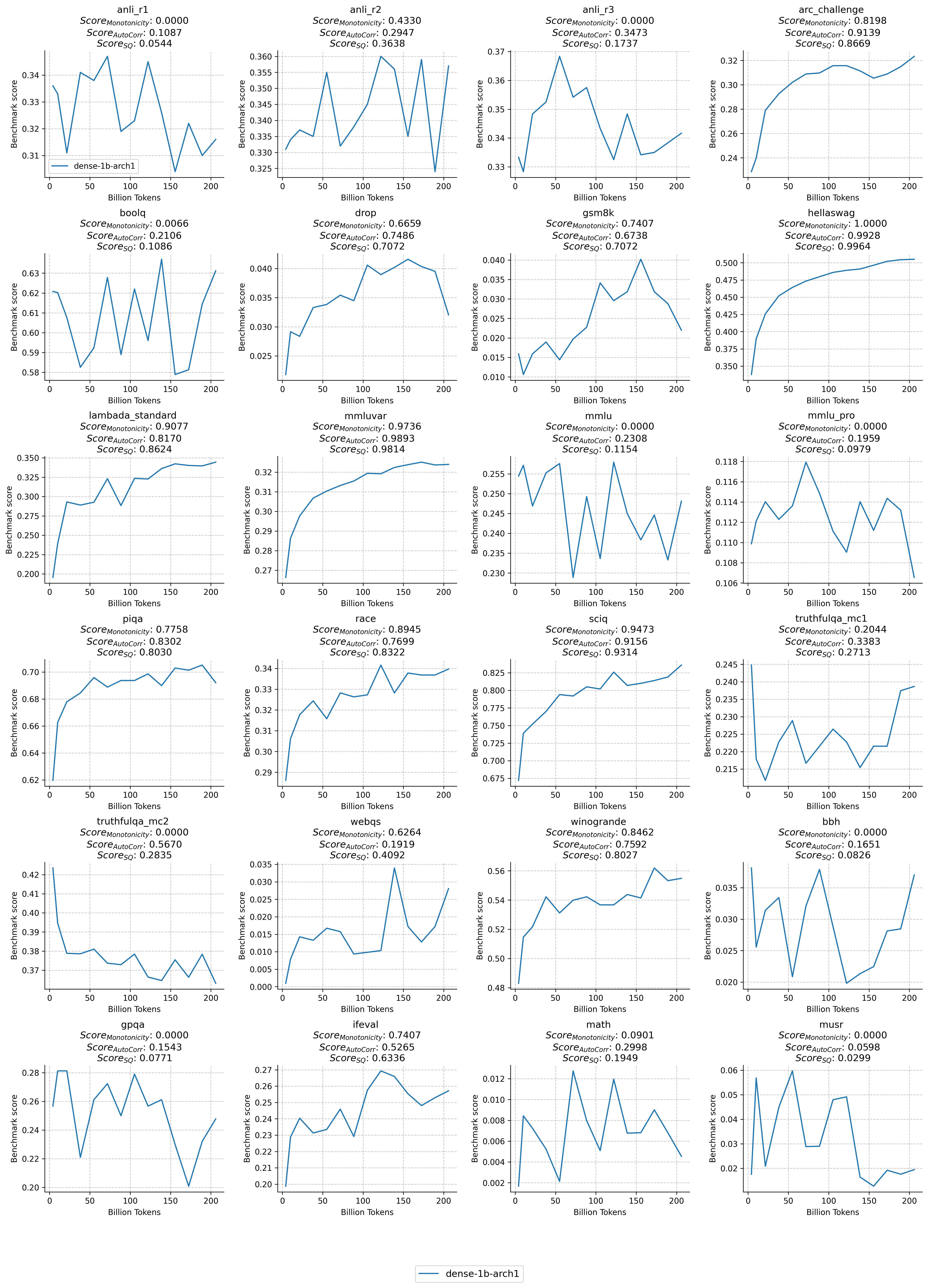}
    \caption{Signal quality metric calculated for different benchmarks using 1B model. (SC corresponds to $\text{Score}_{\text{Monotonicity}}$ in equation \ref{eq: monotonicity}, AC corresponds to $\text{Score}_{\text{AutoCorr}}$ in equation \ref{eq: autocorr} and SQ corresponds to $\text{Score}_{\text{SQ}}$ in equation \ref{eq: score_sq} with $\beta_{\text{Mono}} = \beta_{\text{AutoCorr}} = 0.5$.}
    \label{fig:mmlu-comparison-all-sq}
\end{figure}
\newpage

\subsection[ScoreRC Calculation examples]{$Score_{RC}$ Calculation examples}

\begin{figure}[htbp]
    \centering
    \includegraphics[width=1\textwidth]{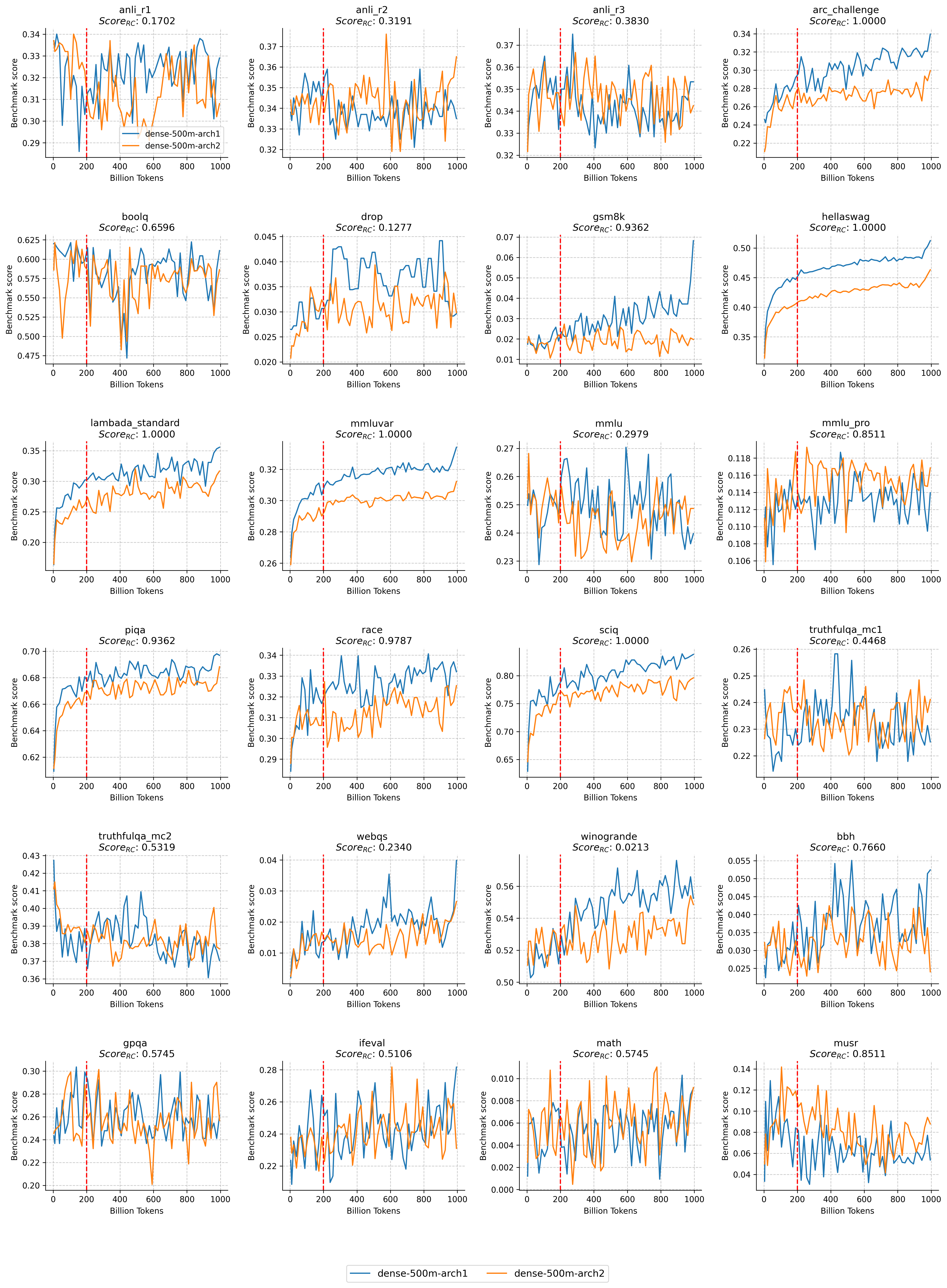}
    \caption{Ranking Consistency metric calculated for different benchmarks using 0.5B model.}
    \label{fig:mmlu-comparison-all-rc}
\end{figure}

\newpage
\subsection[ScoreCS Calculation examples]{$Score_{CS}$ Calculation examples}

\begin{figure}[htbp]
    \centering
     \includegraphics[width=1\textwidth]{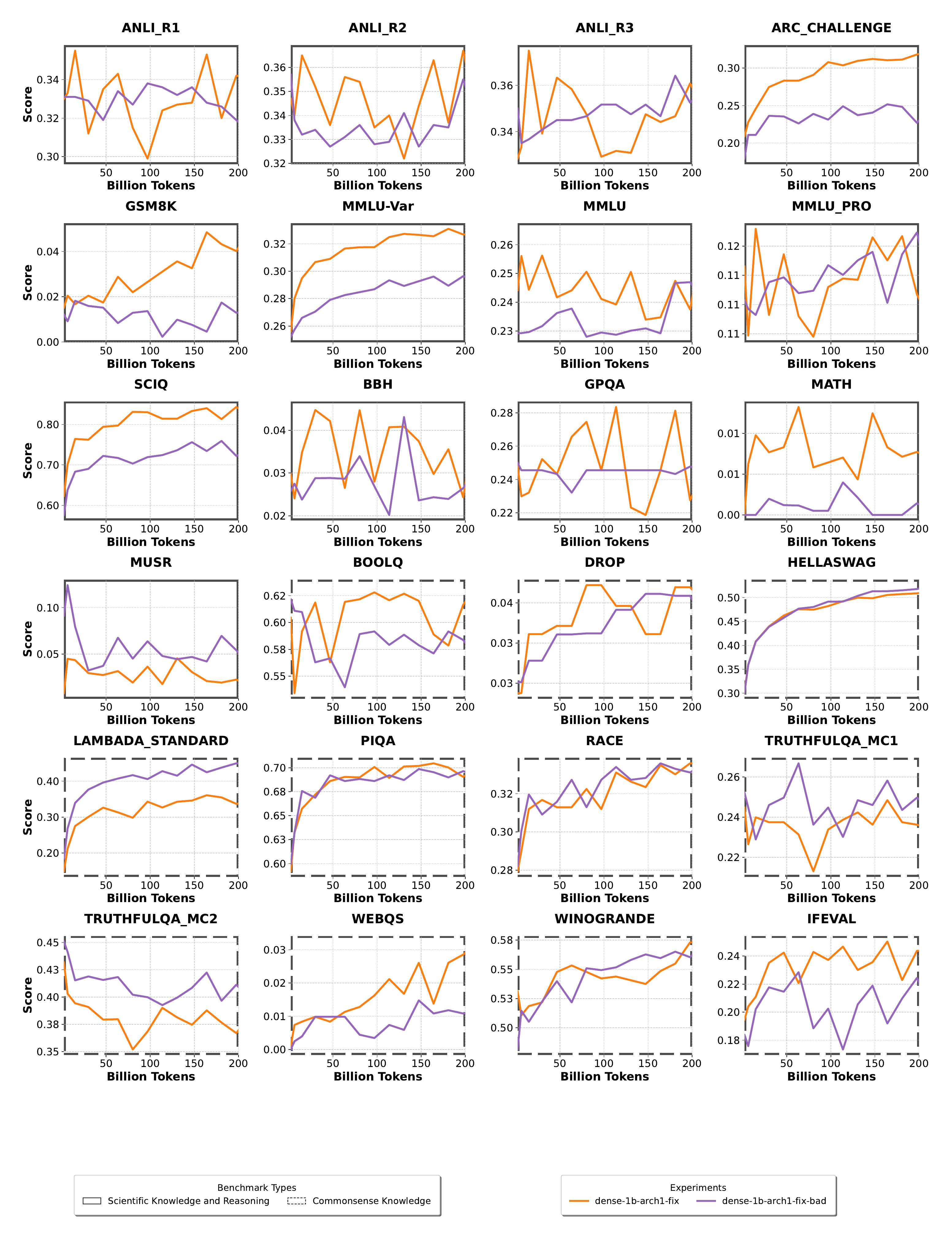}
    \caption{Compliance to scientific knowledge domains metric calculated for different benchmarks using two 1B models, one trained with a web-only data mixture and the other one trained with a rich scientific data mixture. For all the benchmarks, we use acc\_norm as a metric except for TruthfulQA, we use MC1 and MC2.}
    \label{fig:mmlu-comparison-all-cs}
\end{figure}

\subsection{MMLU-var Score Calculation Example} 
\paragraph{Calculating Signal Quality}After Calculating the monotonicity score using Equation~\eqref{eq: monotonicity} and the Autocorrelation Strength using Equation~\eqref{eq: autocorr}, we compute the Signal quality score using Equation~\eqref{eq: score_sq}, where $\beta_{Monotonicity}=0.5$ and $\beta_{AutoCorr}=0.5$. The results for the 3 models are given in \reftab{tab:signal_quality}:

\begin{table}[H]
   \centering
   \begin{tabular}{l|ccc}
       \hline
       \textbf{Model} & \textbf{Score$_\text{Monotonicity}$} & \textbf{Score$_\text{AutoCorr}$} & \textbf{Score$_\text{SQ}$} \\
       \hline
       dense-500m-arch1 & 0.9387 & 0.9324 & 0.9354 \\
       dense-1b-arch1 & 0.9736 & 0.9893 & 0.9814 \\
       dense-3b-arch1 & 0.9516 & 0.9685 & 0.9601 \\
       \hline
   \end{tabular}
   \caption{Detailed Signal Quality scores across the different models}
   \label{tab:signal_quality}
\end{table}

Then, we average over the different model sizes to get the final Signal Quality score: 

\[
    \textbf{Score$_\text{SQ}$} = \frac{0.9354 + 0.9814 + 0.9601}{3}=0.9590
\]

\paragraph{Calculating Ranking Consistency} We compute the ranking at 200 billion tokens using Equation~\eqref{eq:baseline_ranking} then, the ranking consistency until 1 trillion tokens using Equation~\eqref{eq:ranking_consistency} for each model size $s \in \{0.5B, 1B, 3B\}$.  The results are reported in \reftab{tab:ranking_consistency}:

\begin{table}[H]
   \centering
   \begin{tabular}{l|c}
       \hline
       \textbf{Models} & \textbf{Score$_\text{RC}$} \\
       \hline
       (dense-0.5b-arch1, dense-0.5b-arch2) & 1.0 \\
       (dense-1b-arch1, dense-1b-arch2) & 0.5106 \\
       (dense-3b-arch1, dense-3b-arch2) & 1.0 \\
       \hline
   \end{tabular}
   \caption{Detailed Ranking Consistency scores across different model sizes}
   \label{tab:ranking_consistency}
\end{table}

After that we average on model sizes: 

\[
    \textbf{Score$_\text{RC}$} = \frac{1.0 + 0.5106 + 1.0}{3}=0.8368
\]

\paragraph{Calculating Compliance Score} We compute the compliance score using Equation~\eqref{eq:compliance_score} between the two 1B models with the same architecture (Arch-1), but one was trained with a rich scientific knowledge-focused datamixture and the other one was trained using a web-only datamixture. We obtained for MMLUvar the result: 0.384

\paragraph{Calculating the total score} Using Equation~\ref{eq:total_score}, we compute our metric's total score by setting $\alpha_{SQ}=0.4$, $\alpha_{RC}=0.2$, and $\alpha_{CS}=0.4$. 

\[
    \textbf{Score} = 0.5 \times 0.9590 + 0.1 \times 0.8369 + 0.4 \times 0.384 = 0.7167
\]

\section{Check procedures}

\subsection{Scientific alignement check}
\label{scientific-alignement check}

To ensure that submitted benchmarks adhere to the expected \textit{scientific knowledge domains}, we implement an initial compliance check using \textit{gpt-4o-2024-08-06} as a zero-shot classifier. This step is designed to automatically filter out benchmarks that rely solely on general language understanding or common-sense reasoning, and to promote benchmarks that demand substantive academic or scientific knowledge.

\textbf{Classification Protocol:} Each submission is evaluated by prompting \textit{gpt-4o-2024-08-06} with a structured instruction that defines two categories:

\begin{itemize}[leftmargin=1.5em]
  \item \textbf{Accept}: for tasks requiring domain-specific, scientific, or expert-level reasoning.
  \item \textbf{Reject}: for tasks relying only on general language skills, common sense, or trivia.
\end{itemize}

The complete classification prompt is provided below:

\newpage

\thispagestyle{empty}
\begin{tcolorbox}[title=Scientific scoring prompt, colback=white, colframe=black, 
                  left=1pt, right=1pt, top=1pt, bottom=1pt, boxsep=1pt, fontupper=\scriptsize]
\section*{Classification Instructions}

You are tasked with classifying each of the following question-answer pairs into one of two groups:

\subsection*{Accept}

Classify as \textbf{Accept} if the question requires \textbf{domain-specific knowledge}, \textbf{scientific understanding}, \textbf{academic expertise}, or \textbf{professional reasoning}. This includes but is not limited to:

\subsubsection*{Scientific \& Technical Fields}
\begin{itemize}
    \item Biology
    \item Chemistry
    \item Physics
    \item Mathematics
    \item Computer Science
    \item Engineering (all disciplines)
    \item Medicine and Health Sciences
    \item Neuroscience
    \item Pharmacology
    \item Veterinary Science
    \item Environmental Science
    \item Earth Science / Astronomy
    \item Statistics \& Data Science
\end{itemize}

\subsubsection*{Professional \& Applied Domains}
\begin{itemize}
    \item Law
    \item Business (e.g., Finance, Accounting, Marketing)
    \item Economics
    \item Political Science
    \item Education
    \item Sociology
    \item Anthropology
    \item Linguistics
    \item Communications / Media Studies
    \item Library \& Information Science
    \item Social Work
    \item Public Policy
    \item Nursing / Allied Health
    \item Architecture / Urban Planning
    \item Agriculture / Food Science
\end{itemize}

\subsubsection*{Humanities with Reasoning Requirements}
\begin{itemize}
    \item History
    \item Philosophy (e.g., logic, ethics, epistemology)
    \item Theology / Religious Studies
    \item Art History
    \item Literary Theory
\end{itemize}

Any question that falls under these or similar knowledge-intensive domains should be marked \textbf{Accept}, even if it's not explicitly listed.

\subsection*{Reject}

Classify as \textbf{Reject} only if the question is based on:
\begin{itemize}
    \item General language understanding
    \item Common sense or cultural knowledge
    \item Vocabulary, grammar, spelling, or idioms
    \item Word analogies or word associations
    \item Trivia or factoids that don't require reasoning
    \item Sentiment, emotion, or tone recognition
    \item Simple reading comprehension without technical content
    \item NLP-specific tasks (e.g., joke detection, paraphrasing, summarization)
\end{itemize}

These do not require deep subject-matter knowledge and should be marked \textbf{Reject}.
\end{tcolorbox}

\textbf{Thresholding and Manual Review:} For each benchmark, we compute the proportion of questions classified as \textbf{Accept}. If this classification accuracy exceeds \textbf{80\%}, the benchmark is considered \textbf{automatically compliant} and advances to the next evaluation stage.

If the \textbf{Accept accuracy is below 80\%}, the submission is flagged for \textbf{manual review} by domain experts. This ensures that potentially valuable but misclassified benchmarks are not prematurely excluded.

\textbf{Observed Performance on Example Benchmarks:} To calibrate the prompt, we tested the classification protocol on existing benchmarks:

\begin{itemize}[leftmargin=1.5em]
  \item \textbf{HellaSwag} (focused on common-sense reasoning): \textbf{10\% Accept}, indicating low compliance with the scientific knowledge requirement.
  \item \textbf{MMLU}: \textbf{100\% Accept}, demonstrating full alignment with domain expectations.
\end{itemize}

As shown in\reftab{tab:classif_gpt4}, the benchmarks are well classified, where the scientific ones get a high accuracy whereas the rest obtains a lower score (HellaSwag, TruthfulQA, WinoGrande). This automated classifier serves as an efficient preliminary filter while still allowing judgment in edge cases.

\begin{table}[htbp]
\centering
\begin{tabular}{c|c}
\hline
\textbf{Benchmark} & \textbf{Gpt4-o classification accuracy} \\
\hline
MMLU & 100 \\
HellaSwag & 10 \\ 
ARC-Easy & 93 \\
SciQ & 87 \\
TruthfulQA & 27  \\
WinoGrande & 7 \\
MuSR & 34 \\
\hline
\end{tabular}
\caption{Gpt4-o classification of the common benchmarks using the detailed prompt}
\label{tab:classif_gpt4}
\end{table}

\subsection{Leakage check}
\label{leakeage-check}

We provide an additional leakage check to make sure the expected answer is not semantically present within the context, by doing a simple string match between elements in the answer with the context. We calculate the leakage rate for SCiQ\cite{SciQ} benchmark using the snippet below and obtained a score of 0.82, which might explain the quality of signal of that benchmark for small models.

Below is an example of the test set of SCiQ benchmark which contains answer leakage within the passed context.

\begin{tcolorbox}[title=Question sample from SCiQ, colback=white, colframe=black]
Oxidants and Reductants\\
Compounds that are capable of accepting electrons, such as O$_2$ or F$_2$, are called \colorbox{yellow}{\textbf{oxidants}} (or oxidizing agents) because they can oxidize other compounds. In the process of accepting electrons, an oxidant is reduced. Compounds that are capable of donating electrons, such as sodium metal or cyclohexane (C$_6$H$_{12}$), are called reductants (or reducing agents) because they can cause the reduction of another compound. In the process of donating electrons, a reductant is oxidized. These relationships are summarized in Equation 3.30: Equation 3.30 Saylor URL: http://www.saylor.org/books.

\textbf{Question:} Compounds that are capable of accepting electrons, such as O$_2$ or F$_2$, are called what?

\textbf{Answer:}
\end{tcolorbox}

\begin{tcolorbox}[title=Answers sample from previous sample, colback=white, colframe=black]
\begin{itemize}
\item residues
\item antioxidants
\item Oxygen
\item \colorbox{yellow}{\textbf{oxidants}}
\end{itemize}
\end{tcolorbox}

We will conduct similar checks for all proposed benchmarks and provide a score for each of them when relevant. For other benchmarks such as MMLU, this score is not relevant as answers are formatted in MCQ format:

\begin{tcolorbox}[title=Question and Answer sample from MMLU, colback=white, colframe=black]
Find the degree for the given field extension Q(sqrt(2), sqrt(3), sqrt(18)) over Q.

A. 0

B. 4

C. 2

D. 6

Answer:
\end{tcolorbox}

Since the expected answer from the model (choice from A, B, C and D) is not within the context, this check is not relevant for MCQ-style benchmarks.

\newpage

\section{Global scores}
The table presents a comprehensive ranking of 25 benchmarks based on their performance across three metrics that we proposed. MMLU-var ranks highest with a total score of 0.717, followed by ARC-Easy and SciQ in the second and third positions with total scores of 0.655 and 0.627 respectively. Interestingly, HellaSwag achieved very high scores on the first two metrics, as it is well adapted to early training assessment. However, it received a score of 0 on the third metric as it is not designed to assess scientific knowledge.
Note that several tasks known to be relevant to scientific knowledge, such as Math and MMLU-Pro, received relatively poor scores. This is principally due to the fact that their low score in the SC metric is correlated with their low score in the SQ metric, as they are noisy and fail to provide a clear signal at this stage of training, making them unable to discriminate effectively between the web-only model and the knowledge-oriented model.
\label{global-score-all}

\begin{table}[htbp]
\centering
%\rowcolors{2}{gray!10}{white}
\begin{tabular}{>{\bfseries} l c c c c}
\toprule
 \textbf{Benchmark} & \textbf{SQ (50\%)} & \textbf{RC (10\%)} & \textbf{SC (40\%)} & \textbf{Total Score} \\
 \midrule
 \rowcolor{gray!30}
 MMLU-var\cite{muennighoff2024olmoe} & 0.959 & 0.837 & 0.384 & 0.717 \\
 \rowcolor{gray!30}
 ARC-Easy\cite{allenai:arc} & 0.832 & 0.822 & 0.394 & 0.655 \\
 \rowcolor{gray!30}
 SciQ\cite{SciQ} & 0.846 & 0.772 & 0.316 & 0.627 \\
 HellaSwag\cite{zellers2019hellaswag} & 0.992 & 0.936 & 0.000 & 0.590 \\
 \rowcolor{gray!30}
 GSM8K\cite{cobbe2021training} & 0.655 & 0.915 & 0.369 & 0.566 \\
 LAMBADA\_Standard\cite{paperno-EtAl:2016:P16-1} & 0.894 & 0.752 & 0.000 & 0.522 \\
 PIQA\cite{Bisk2020} & 0.842 & 0.690 & 0.013 & 0.495 \\
 IFEval\cite{zhou2023instructionfollowingevaluationlargelanguage} & 0.532 & 0.562 & 0.353 & 0.463 \\
 RACE\cite{lai-etal-2017-race} & 0.677 & 0.877 & 0.000 & 0.426 \\
 WebQuestions\cite{berant-etal-2013-semantic} & 0.477 & 0.573 & 0.256 & 0.398 \\
 \rowcolor{gray!30}
 MATH\cite{hendrycks2021measuring} & 0.266 & 0.672 & 0.494 & 0.398 \\
 WinoGrande\cite{sakaguchi2021winogrande} & 0.590 & 0.588 & 0.022 & 0.363 \\
 \rowcolor{gray!30}
 MMLU\cite{hendryckstest2021} & 0.285 & 0.574 & 0.400 & 0.360 \\
 \rowcolor{gray!30}
 BBH\cite{suzgun2022challenging} & 0.352 & 0.675 & 0.283 & 0.357 \\
 DROP\cite{Dua2019DROP} & 0.573 & 0.386 & 0.059 & 0.349 \\
 BoolQ\cite{clark2019boolq} & 0.327 & 0.771 & 0.188 & 0.316 \\
 ANLI\_r2\cite{nie2019adversarial} & 0.296 & 0.433 & 0.241 & 0.288 \\
 \rowcolor{gray!30}
 MMLU\_Pro\cite{wang2024mmlu} & 0.342 & 0.716 & 0.000 & 0.243 \\
 \rowcolor{gray!30}
 GPQA\cite{rein2024gpqa} & 0.227 & 0.402 & 0.049 & 0.173 \\
 \rowcolor{gray!30}
 ANLI\_r3\cite{nie2019adversarial} & 0.224 & 0.425 & 0.000 & 0.154 \\
 TruthfulQA\_mc2\cite{lin2021truthfulqa} & 0.180 & 0.625 & 0.000 & 0.152 \\
 TruthfulQA\_mc1\cite{lin2021truthfulqa} & 0.226 & 0.355 & 0.000 & 0.149 \\
 \rowcolor{gray!30}
 MuSR\cite{sprague2310musr} & 0.116 & 0.713 & 0.000 & 0.129 \\
 \rowcolor{gray!30}
 ANLI\_r1\cite{nie2019adversarial} & 0.148 & 0.319 & 0.000 & 0.106 \\
\bottomrule
\end{tabular}
\caption{Benchmark Performance Comparison by Rank. \textbf{SQ} = Signal Quality (50\%), \textbf{RC} = Ranking Consistency (10\%), \textbf{SC} = Scientific Knowledge Compliance (40\%).\textbf{Gray-shaded rows} indicate benchmarks classified as targeting scientific knowledge \& reasoning. Other tasks are related to commonsense knowledge}
\label{tab:benchmark_comparison}
\end{table}

\section{Tracking Validation loss}
\label{val-loss}
\begin{figure}[t!]
    \centering
    \includegraphics[width=\textwidth]{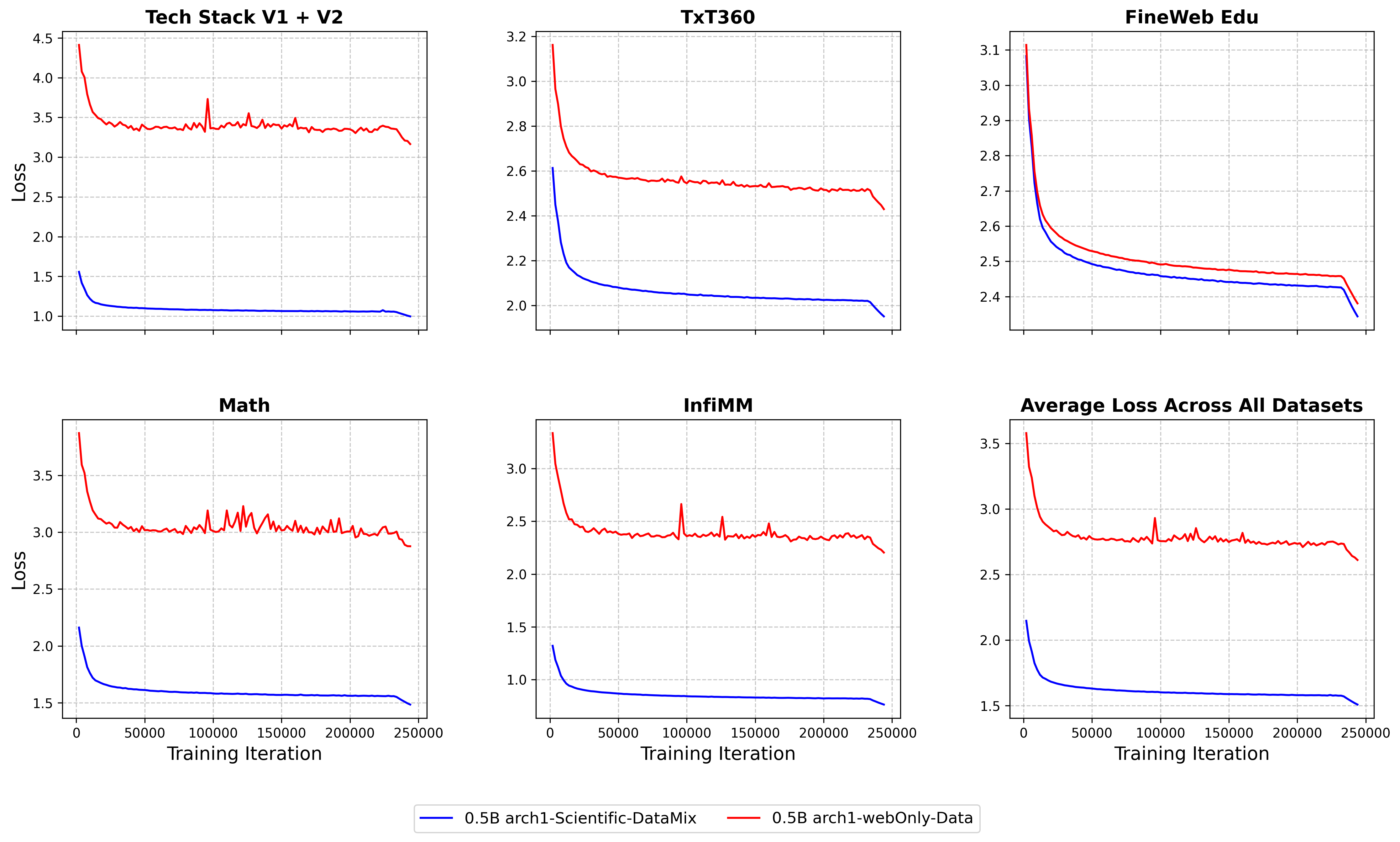}
    \caption{Validation loss across different datasets for two model variants: Scientific-DataMix (blue) and webOnly-Data (red).}
    \label{fig:loss}
\end{figure}

Tracking validation loss during model training is a standard practice in machine learning, with LLM scaling laws using predicted loss as a proxy for model capability. Figure~\ref{fig:loss} shows validation loss across training iterations for two model variants trained on different data mixtures: Web-Only and Scientific-Mix.

Our analysis of these loss curves reveals several important findings:

\begin{enumerate}
    \item \textbf{Domain transfer gap:} While FineWebEdu (a subset of the training data for Web-Only) shows a small gap between variants, all other datasets exhibit substantially larger gaps, often exceeding the total reduction in loss throughout training, indicating significant out-of-distribution effects for the WebOnly model.
    
    \item \textbf{Cross-dataset comparison challenges:} Each dataset produces distinct numerical loss ranges due to inherent characteristics, making direct comparison across domains difficult. Lower loss on one dataset does not necessarily translate to higher capability on corresponding benchmarks.
    
    \item \textbf{Loss interpretation limitations:} The small numerical differences in loss values throughout training are difficult to translate into meaningful capability improvements, particularly when averaging across datasets with different baseline values.
    
    \item \textbf{Early training ambiguity:} Despite Scientific-Mix eventually showing lower loss on FineWebEdu compared to WebOnly, both variants exhibit similar loss trajectories in early training stages, making it challenging to distinguish performance differences during this critical period.
\end{enumerate}

These observations highlight the limitations of using raw validation loss as the sole metric for evaluating model capability across different data mixtures and domains, especially during the early phase of training.

\end{document}